\documentclass[11pt]{article}
\usepackage[final]{acl}

\usepackage{newtxtext,newtxmath}

\usepackage{latexsym}

\usepackage{amsmath}
\usepackage{amsfonts}

\usepackage{amssymb}
\usepackage[disable]{todonotes}
\usepackage[T1]{fontenc}
\usepackage[utf8]{inputenc}

\usepackage{microtype}

\usepackage[safe]{tipa}

\usepackage{inconsolata}

\usepackage{graphicx}

\usepackage{caption}
\usepackage{subcaption}

\usepackage{booktabs}
\usepackage{multirow}
\usepackage{adjustbox}

\title{

Measuring cross-language intelligibility between Romance languages 
with computational tools
}

\author{
Liviu P Dinu$^{\spadesuit,\heartsuit} \mbox{   }\mbox{   }$ Ana Sabina Uban$^{\spadesuit,\heartsuit} \mbox{   }\mbox{   }$ 
\textbf{Bogdan Iordache}$^{\spadesuit,\heartsuit} \mbox{   }\mbox{   }$ \\
\textbf{Anca Dinu}$^{\spadesuit,\heartsuit} \mbox{   }\mbox{   }$
\textbf{ Simona Georgescu}$^{\clubsuit,\heartsuit} \mbox{   }\mbox{   }$ \\
  University of Bucharest,  $^\spadesuit$ Faculty of Mathematics and Computer Science,
   \\
  $^\clubsuit$Faculty of Foreign Languages and Literatures, $^\heartsuit$HLT Research Center \\
  {\tt  \{ldinu, auban\}@fmi.unibuc.ro, ioan.iordache@s.unibuc.ro}, \\ 
 \vspace{-2mm}
  {\tt anca.dinu, simona.georgescu@lls.unibuc.ro}  \\ }

\begin{document}
\maketitle
\begin{abstract}
We present an analysis of mutual intelligibility in related languages applied for languages in the Romance family. We introduce a novel computational metric for estimating intelligibility based on lexical similarity using surface and semantic similarity of related words, and use it to measure mutual intelligibility for the five main Romance languages (French, Italian, Portuguese, Spanish, and Romanian), and compare results using both the orthographic and phonetic forms of words as well as different parallel corpora and vectorial models of word meaning representation. The obtained intelligibility scores confirm intuitions related to intelligibility asymmetry across languages and significantly correlate with results of cloze tests in human experiments.
\end{abstract}

\section{Introduction}

\subsection{Defining intelligibility}

By "cross-language intelligibility" we refer to a speaker's ability to understand an unknown foreign language based on already acquired language skills and resources, mainly from their mother tongue. This ability is also known in the literature as "intercomprehension” (Reinheimer-Ripeanu 2001; Doyé 2005), "receptive multilingualism” (Braunmüller and Zeevaert 2001; Goosekens 2019)\todo{Toate citarile astea ar trebui puse cu \\cite{} si referinte din custom.bib}, or "mutual intelligibility” (Ciobanu and Dinu 2014; Goosekens et al. 2018). 
Although early studies in the field of intelligibility considered reciprocity to be a basic feature of intercomprehension (Voegelin and Harris 1951; Hickerson, Turner, and Hickerson 1952; Pierce 1952; Casad 1974),  empirical evidence increasingly supports that mutual (symetrical) intelligibility is rare even between two language pairs (Goosekens 2024). Therefore, in more recent studies, intelligibility is described either as nonreciprocal (Kluge 2007) or, more frequently, as asymmetric (Gooskens and Van Bezooijen 2006; Gooskens 2007; Gooskens, Van Heuven, et al. 2010). We use the term cross-language intelligibility (Gooskens and Van Heuven 2017) precisely because, as a result of this asymmetry, the levels of comprehension of$L_1$ for speakers of $L_2$ may differ from the intelligibility of $L_2$ for speakers of $L_1$. While mutual intelligibility represents an average of the two levels of comprehension, cross-language (or nonreciprocal) intelligibility captures the asymmetries as they are reflected in the data.

Our aim in this paper is to conduct a quantitative (and qualitative?) analysis of cross-language intelligibility between any two Romance languages from the core group (Romanian, Italian, French, Spanish and Portuguese) through computational methods, in order to identify asymmetries between language pairs and to offer a potential explanation for them.

From the outset, we need to distinguish between \textit{inherent intelligibility}, i.e., based on linguistic features that are common or similar in $L_1$ and $L_2$, and \textit{acquired intelligibility}, which relies on a certain degree of prior learning and normally involves less closely related languages (Gooskens, 2019; Stenger, Jágrová, and Avgustinova 2020). This study focuses on inherent intelligibility, as we do not examine actual speakers or incorporate extra-linguistic information. Instead, our goal is to establish a benchmark: using a set of authentic linguistic data, we investigate the extent to which an average speaker of a Romance language $L_1$ can understand another Romance language $L_2$ in the absence of prior exposure.

Although we are aware of the existence of multiple extra-linguistic factors involved in cross-language intelligibility like exposure,  attitudes, or plurilingual resources,  we focus in this paper on  linguistic factors only (including lexical and the semantical layers), since, unlike extralinguistic factors, they are a quantifiable component of language.

\subsection{Motivation}

Research on intelligibility offers crucial insights into the human language faculty and its constraints. Its findings help us better understand the mechanisms involved in human language processing, allowing us to identify how much a language can diverge from a listener’s native tongue before it becomes unintelligible (cf. Goosekens 2024).
Language intelligibility is actually mentioned as a vivid problem in the report published in 2007 at the European Commission by the High Level Group on Multilingualism (HLGM), which emphasizes “a lack of knowledge about mutual intelligibility between closely related languages in Europe and the lack of knowledge about the possibilities for communicating through receptive multilingualism, i.e., where speakers of closely related languages each speak their own language”. Several projects have been carried out to promote this type of communication (EuRom 4 [www.up.univ-mrs.fr/delic/Eurom4], LaLiTa [www.ciid.it/lalita], EuroComRom [Kleine and Stegmann 2000] etc.). Nonetheless, to effectively improve mutual understanding, it is essential to assess how intelligible closely related languages are to one another and to identify the linguistic factors that affect this intelligibility.

\subsection{Previous work}

The causes of asymmetry have generally been explained as a kind of imbalance between languages A and B with respect to the factors influencing intelligibility, whether extra-linguistic (exposure, plurilingual resources, attitude, orthographic knowledge, etc.) or linguistic (lexicon, pronunciation, morphology, syntax, paralinguistics). Most concrete studies on asymmetry focus on pairs of Germanic languages (mostly Danish–Swedish, cf. Gooskens and van Bezooijen 2013; Schüppert, Hilton, and Gooskens 2015), Slavic languages (cf. Golubovic and Gooskens 2015; Fischer and Stenger 2016; Renslow 2018), Chinese dialects (Chaoju and van Heuven), or lesser-known languages, such as Indigenous American languages (see the pioneering research of Voegelin and Harris 1951; Hickerson, Turner, and Hickerson 1952) or languages from Vanuatu (Gooskens and Schneider 2019).

For Romance languages, as far as we know, there are no systematic studies; research on this linguistic family exists mostly as a component of broader investigations (Heeringa et al. 2013; Schüppert et al. 2017), with analyses largely based on cloze tests (cf. Gooskens 2024). Yet this particular language family is especially interesting in this regard, precisely because it provides the largest and most reliable diachronic data, which would allow for identifying historical, social, and cognitive explanations for asymmetry. It could serve as an important basis for study, enabling the refinement of methods and tools so as to constitute a benchmark for future studies in other families.
Moreover, unlike in the Slavic and Germanic families, the source language of Romance varieties —Latin—is well documented, offering crucial insights into the causes of diversification and divergence, including factors such as Latin polysemy, word frequency, connotations, and synonym preferences.

Moreover, unlike studies that focus on intelligibility and are limited to the strict observation of these phenomena, we aim to open up broader theoretical questions.

One of the issues for which we seek to provide an initial step toward a solution is the following: even though the Romance languages are closely related, not only do degrees of similarity between any two Romance languages vary from one author to another, but their classification is also controversial. For example, \citet{mcmahon2003finding} report two different results for the classification of Romanian within the Romance family, either marginal or more integrated within the group. Our approach would provide evidence either sustaining or invalidating the theory of lateral areas \cite{bartoli1925introduzione}, enabling the distinction between inherited affinity and gained similarity, due to language contact.

\todo{Anca: ar trebui aici inclus si ceva despre translation scores, dupa cum a cerut un referent: what about results on using translation scores to rebuild family trees? Translation seems like a very good way of modelling intelligibility under the assumption that the closer the easier to translate. This goes back to Philipp Koehn’s work, it is not recent.}

\subsection{Contribution}

Many of these ideas were long confined to the theoretical domain because appropriate tools for gathering and analyzing data were unavailable (Goosekens 2024). Most studies on mutual intelligibility have relied on small, hand-selected datasets for the languages under investigation The limitation of this approach is that such datasets do not adequately reflect the full scale and diversity of a language. To obtain reliable estimates of mutual intelligibility between two languages, it is necessary to compare a larger and more representative subset of their vocabulary (Renslow 2018). Historical linguists often use word lists, such as the Swadesh list (Swadesh 1971) and the Leipzig-Jakarta list (Tadmor 2009), to assess the degree of chronological divergence between languages. However, when the goal is to predict or explain intelligibility among contemporary languages, it may be more effective to base the analysis on a representative or random sample of each language’s vocabulary, and – most important – vocabulary in context.

This is the first aspect in which our study makes a significant contribution, as the corpus we use comprises two components. For the first level – lists of words – , we exploit the RoBoCoP database, a complete storage of cognates and borrowings between any two Romance languages \cite{dinu2023robocop}. By providing all the Romance cognate pairs identified between any two of the five main Romance languages we are dealing with, RoBoCoP enables an exhaustive perspective on the lexical relationship between the Romance languages.
However, it is not sufficient to track words outside of their contexts (as shown by Gooskens 2024, who obtained much better results when words were included in phrases or sentences). Therefore, the cognates and loanwords automatically extracted from the RoBoCoP database are analyzed in context, across two types of texts (see Section Dictionaries and Corpora).

The second significant innovation of our study is methodological in nature. Until now, intelligibility has mostly been studied at the lexical level, focusing on the recognition of cognates. When listening to a closely related language, listeners are often able to recognize cognates because they are similar to words in their native language. Cognates can be defined as historically related words in the vocabularies of two related languages, i.e. word pairs that can be traced back to the same lexical ancestor, be they inherited (e.g. Ro. \textit{ochi} 'eye' – It. \textit{occhio} - Fr. \textit{oeil}, Es. \textit{ojo}, Pt. \textit{olho}, from Lat. \textit{oculus}), or borrowed (e.f. Ro. \textit{zahar} 'sugar' – Fr. \textit{sucre} – Sp. \textit{azúcar}, etc. from Ar. \textit{sukkar}). Studies have generally started from the premise that lexical distance shows a strong correlation with intelligibility scores: a higher proportion of non-cognates in the target language reduces comprehensibility, whereas a greater presence of cognates facilitates understanding. Nonetheless, it goes without saying that cognate relationships are only helpful if listeners are able to identify words as cognates (see Goosekens 2024). Inherited cognates, even in closely related languages, may have undergone such significant changes in form that listeners will no longer be able to match them with the counterparts in their own language (Ro. \textit{genunchi} 'knee' – Es. \textit{hinojo}, or It. \textit{madre} - Fr. \textit{mère}).

Accordingly, we highlight two key considerations that will guide our analysis and that should be taken into account in future approaches to intelligibility.
Firstly, a substantial lexical overlap between two languages does not, by itself, guarantee high intelligibility, since the pronunciation of cognates can differ to such an extent that listeners may fail to recognize them as related. Put differently, lexical overlap constitutes a necessary, but not sufficient, condition (Goosekens 2024).
Secondly, Otwinowska and Kasztelanic 2011 show that the cognate facilitation effect is influenced not only by the level of formal similarity between cognates, but also by the transparency of their lexical meaning.

 In light of this, we introduce a method that takes multiple factors into account: (1) the lexical overlap between languages (i.e., the number of cognates and borrowings between any two Romance languages); (2) the phonetic level, comprising two components: pronunciation and orthography; (3) the degree of semantic proximity, which is the most difficult aspect to capture in a quantitative measurement, to the best of our knowledge never approached before. In the following section, we explain how we will address each point.

\section{Methodology}
\label{sec:methods}

(1) For the first point, we benefit from using the largest database, RoBoCoP, which we apply to texts. Given that not all pairs of words are alike (e.g. Ro. \textit{om} ‘man’ is formally more distant from Sp. \textit{hombre} than Ro. \textit{luna} ‘moon’ is from Sp. \textit{luna}), we apply a measurement method to quantify the similarity of word forms, and we will quantify these results.

(2) With regard to the second point, we proceed from the premise that identifying orthographic correspondences between languages can increase the intelligibility (Fisher and Stenger 2016), while pronunciation may pose a barrier to cross-linguistic intelligibility. For example, if a speaker of any Romance language sees the Fr. word \textit{temps} written down, they are very likely to be able to relate it to the cognate in their own language (It. \textit{tempo}, Es. \textit{tiempo}, Pt. \textit{tempo}, Ro. \textit{timp}), but there is little chance that they will be able to identify the word when they hear it pronounced [tã].

(3) As for the third aspect, we must bear in mind that cognates, in fairly many cases, have diverged in meaning (vezi articolul nostru Friend or Foe). Therefore, the traditional approach, in which lexical distance between two languages is calculated simply by computing the percentage of non-cognates is far from being sufficient: e.g. Ro. \textit{lume} ‘world’ has diverged semantically from Lat. \textit{lumen} ‘light’, which hinders its identification as a cognate of Sp. \textit{lumbre} ‘light, fire’; or, in another case, both Ro. \textit{larg} ‘wide’ and Es. \textit{largo} ‘long’ have diverged from Lat. \textit{largus} ‘abundant’. Many such cognate pairs become deceptive cognates, or, in terms of \citet{dominguez2002false}, semantic false friends. While regular non-cognates generally impede intelligibility, false friends can actually create greater difficulties because they have the potential to mislead the listener: since listeners may not recognize that these words carry different meanings, they are less likely to rely on contextual cues to infer their meaning, compared with words whose forms are clearly distinct. \cite{gooskens2024mutual}.
Therefore, our study covers the semantic level. To measure semantic similarity, we determine the degree of meaning divergence in all cognate pairs, for all pairs of Romance languages, and, through an appropriate mathematical formula, we combine this divergence with the degree of closeness of the forms, determined previously. As a testing and validation method, the computer also provides the lexeme that is semantically closest to the target word. (e.g. Es. \textit{pariente} 'relative', although cognate of Ro. \textit{parinte}, does not share its meaning of 'parent'; therefore, the algorithm will also provide the semantically closest Romanian word for Es. \textit{pariente}, i.e. \textit{ruda}, and vice versa - Es. \textit{padre}).

\subsection{Dictionaries and Corpora}
\label{sec:data}

% EU AS SCHIMBA IN DOUA SUBSECTIUNI, SI AS REDENUMI DATA SECTIUNEA MARE. AS PUNE PRIMA SUBSECTIUNE CORPORA, IN CARE SPUNEM CE CORPORA FOLOSIM, SI A DOUA SUBSECTIUNE SA FIE LEGATA DE RELATED WORDS, IN CARE SA INTROUDCUEM PROTOWORD SI SA SPUNEM CU CE DATE AM LUCRAT CA SA LE EXTRAGEM DE ACOLO.

% AS SPUNE CA SUNTEM CONSTIENTI CA REZULTATELE SUNT INFLUENTATE DE CALITATEA CORPUSULUI, CEVA DE FORMA ASTA, CA POATE P ELIMBA VORBITA SUNT ALTE REZULTATE SI CA GRADUL DE APROPIERE DEPINDE DE CALITATEA DATELOR, DE COPUSUL AVUT, DAR, IN LIPSA UNOR CORPUSURI PARALELE, AM FOLOSIT CE AVEM
%

Since we start from the premise that intercomprehension is based on the ability of the average speaker to identify words in the flow of speech, we ground our analysis on authentic acts of speech, not simply lexicographic works. Therefore, we analyze two public corpora for the five Romance languages, RomCro - containing literary texts in different languages, translated in Romance languages and Croatian \citep{mikelenic2024expansion}, and EuroParl - focusing on proceedings of the European Parliament \citep{koehn2005europarl}. 
We extract related word frequencies in corpora used in our metrics based on the two parallel corpora \todo{Anca: rephrase pls. nu se intelege}. 

We perform our analysis on related word pairs extracted from the most comprehensive database of related words in cognate languages up to date, sourced from etymological dictionaries and manually curated, RoBoCoP \cite{dinu2023robocop}. As a source of cognate word pairs, we use the freely available subset ProtoRom \cite{dinu2024verba}, a database of cognate tuples and etymons in the five Romance languages,  with 19,222 entries (tuples with at least 2 cognates). We extract borrowings from the original RoBoCoP database, totaling 46,490 borrowing pairs across Romance language pairs \cite{dinu2024takes}.

In order to identify occurrences in text for our pairs of related words (i.e. cognates and borrowings) for a given language pair, we process parallel sentence examples from the employed datasets as follows: we tokenize the sentences using spaCy \cite{honnibal2020spacy}, remove stop-words, and match each token with a corresponding instance from RoBoCoP, if possible. In order to account for inflections of the dictionary form from RoBoCoP perform normalization including accent removal and stemming, using a Snowball stemmer \cite{porter2001snowball}. For each pair of sentences, we count how many words in one language are related to the words in the other language. We also count for a given related word pair how many times that pair appears in aligned examples, in other words: how many times the related word pair corresponds to a proper translation. Table \ref{tab:corpus-stats} shows these counts.

% AS MAI FACE SI ASTA: EVENTUAL IN APENDIX, AS PUNE DIN RELATED WORDS CATE WUNT BORROWINGS SI CATE COGNATES.

% AS MAI FACE CEVA: AS VREA SA VEDEM SI CAT DE BINE IDENTIFICAM NOI RELATED WORDS IN PROPOZITII. AS SCOATE ALEATOR UN NUMAR DE 1000 DE PROPOZITII, SAU CATE CREDETI CA SUNT STATISTIC RELEVANTE, SI SA FACEM MANUAL RECALLL SI PRECISION PE ELE. SA NU NE PACALIM CUMVA
% LA EUROPARL, AM O MICA INTREBARE DACA NU CUMVA AR TREBUI SA LUAM ACELASSI NUMAR DE PROPOZITII LA FEL CA LA ROMCRO. ASTA E NUMARUL MINIM DAT DE INTRAREA RO IN UE, SI PE ALEA SA LE LUAM. EVENTUAL SA FACEM SI ASA SI SA COMENTAM DACA IDENTIFICAM DIFERENTE?

% SI SA ADAUGAM SI WIKIPEDIA

\subsection{Surface and Semantic Similarity}

% EVENTUAL SA FACEM LA FEL O SUBSECTIUNE MARE INTITULATA WHAT AND HOW DO WE MEASURE SI APOI SA PUNEM CELE 2 SUBSECTIUNI DE ACUM SUB EA?

% SA INCEPEM CU WHAT SI EXPLICAM ASTA DE AICI

The intelligibility computation is twofold: we combine the orthographic/phonetic similarity of word pairs across languages with their semantic similarity.

For the former, we measure string similarity using the normalized Levenshtein distance \cite{levenshtein1966binary} either on the orthographic (after removing accents) or on the phonetic representation (where each phoneme is considered a single unit). We employ the \emph{eSpeak}\footnote{https://github.com/espeak-ng/espeak-ng} library to automatically generate the phonemic representations.   Thus we obtain scores in the interval $[0, 1]$, where $1$ means identical representations.

% BOGDAN, ANA: AICI LA LEVENSTHEIN, AM CALCULAT DISTANTA INTRE FORMELE LEMATIZATE? SAU INTRE CE GASEAM? IN 2014 AM MERS PE FORMELE LEMATIZATE

The latter (i.e. semantic similarity) is computed based on word representations trained on large corpora to represent meaning in context. 

% SA MAI EXPLICAM CE VREM AICI CU SEMANTIC SIMILAIRTY? INTR-UN FOOTNOTE POATE? ADICA SA ZICEM CA ELE ISI MAI POT SCHIMBA SENSUL (UN EXEMPLU), SI CA NOI ASTA CALCUALM DE FAPT?

A metric of semantic similarity is computed using cosine similarity based on FastText word embedding vectors \cite{bojanowski_et_al}. Since we are assessing similarity across languages, we need our embedding spaces to be aligned. For that reason, we use pretrained prealigned static embeddings, based on a large multilingual corpus (Wikipedia) \cite{joulin2018loss}. Since even on a large corpus, such as the ones used to pretrain these embeddings, some of our related words extracted from RoBoCoP may not be represented, but an inflected form may be available, we find for each unrepresented word its closest match in form (via stemming and edit distance) that is present in the pretrained embeddings and we use that vector as our canonical representation. 

%% not really (folosim embeddings preantrenate pe corpus mare) -- For the smaller corpora for which there are no pretrained aligned embedding spaces across languages, we experiment with aligning the obtained spaces in order to obtain a single embedding vector space across languages (one for each corpus) based on a small bilingual dictionary curated manually, following the method in \citet{lample2017unsupervised}. 

We also explore how to integrate contextual representations of these words in order to compute semantic similarities. We use a BERT-like transformer model that was pretrained on a multilingual sentence similarity task for optimizing sentence representations, based on a Sentence-BERT architecture \cite{reimers-2019-sentence-bert} \footnote{https://huggingface.co/sentence-transformers/distiluse-base-multilingual-cased-v2}. Contextual embeddings are computed for each occurrence of the related words across the employed parallel corpora. When computing the similarity score between two related words, in order to capture multiple senses used in the texts, we form embedding clusters for each word in the pair using Affinity Propagation \cite{frey2007clustering}, we identify cluster centers using simple averaging, and the average pairwise cosine distance between cluster centers is the similarity distance we are looking for, previously shown to work well for lexical semantic word representations in the context of computational measures of lexical semantic change \cite{periti2024systematic}.

\begin{table*}[ht]
\centering
\begin{adjustbox}{max width=\textwidth}
\begin{tabular}{llrrrrrrrrrr}
\toprule
\textbf{Dataset} & \textbf{Metric} 
& \textbf{es-fr} & \textbf{es-pt} & \textbf{es-ro} & \textbf{fr-pt} 
& \textbf{fr-ro} & \textbf{it-es} & \textbf{it-fr} 
& \textbf{it-pt} & \textbf{it-ro} & \textbf{pt-ro} \\
\midrule
\multirow{3}{*}{RomCro} 
& Sentences     & 166,738 & 166,738 & 166,738 & 166,738 & 166,738 & 166,738 & 166,738 & 166,738 & 166,738 & 166,738 \\
& Total words     & 3.29M & 3.14M & 3.04M & 3.32M & 3.22M & 3.22M & 3.41M & 3.26M & 3.15M & 3.07M \\
& Related words     & 1.44M & 1.50M & 1.18M & 1.30M & 1.56M & 1.47M & 1.41M & 1.84M & 1.18M & 0.87M \\
& Aligned pairs     & 0.28M & 0.41M & 0.17M & 0.24M & 0.28M & 0.31M & 0.27M & 0.38M & 0.18M & 0.10M \\
\midrule
\multirow{3}{*}{EuroParl} 
& Sentences     & 1,982,990 & 1,933,321 & 387,653 & 1,980,132 & 387,846 & 1,880,982 & 1,943,673 & 1,877,432 & 367,904 & 381,404 \\
& Total words     & 53.92M & 52.01M & 10.04M & 55.82M & 10.50M & 52.37M & 56.68M & 54.21M & 10.32M & 10.11M \\
& Related words     & 21.82M & 20.93M & 4.42M & 18.52M & 6.62M & 21.27M & 19.51M & 32.80M & 4.03M & 2.60M \\
& Aligned pairs     & 4.74M & 5.05M & 0.85M & 3.52M & 1.89M & 4.30M & 3.57M & 7.22M & 0.81M & 0.38M \\
\bottomrule
\end{tabular}
\end{adjustbox}
\caption{Corpus statistics for the two parallel corpora. For each corpus and each dataset we show the total number of parallel sentences, how many word occurrences we have (excluding stop-words) from either language, how many related words occur from either language (those are words that are cognates with or borrowings to/from the other language) and how many occurrences of aligned pairs of related words could be identified (those are pairs of cognates/borrowings for which both words appear at the same time in a parallel sentence example). Frequency values are shown in millions. }
\label{tab:corpus-stats}
\end{table*}

\subsection{Lexical Intelligibility Index}
\label{sec:dindex}

% AICI EVENTUAL SA NUMIM HOW DO WE MEAUSRE? lEXICAL INTELIGIBILITY INDEX.
% EVENTUAL SA ZICEM CA O INTROUDCEM SI SA INCERCAM SI O MOTIVATIE, CEVA?

We introduce a new formula for intelligibility, called $D_{LI}$.
It is computed via the following formula:
\begin{equation}
    D_{LI}=\frac{S_s S_L(2-S_s -S_L)}{1-S_sS_L}
    \label{eq-li}
\end{equation}
where $S_L$ is the formal lexical similarity between two words (i.e. $1$ minus the normalized Levenshtein distance, computed on the orthographic/phonetic representations), and $S_S$ is the semantic similarity between two words (i.e. $1$ minus the cosine distance between static word embeddings or the average cosine distance computed on contextual embedding clusters).

To obtain (1) we start from the assumption that the lexical ineligibility depends both on the surface (orthographic or phonetic) distance between two words and on the semantic divergence between them. So, we propose a linear combination between them and propose $D_{LI}$, a general index of lexical intelligibility, as follows:
\begin{equation}
D_{LI}=\alpha S_S + \beta S_L
\label{eq-li-ab}
\end{equation}
In order to obtain the values $\alpha$ and $\beta$, we made the following presuppositions: when the two words are identical, the formal similarity between them is maximal, i.e. $S_L = 1$, and the intelligibility index is given by their semantic similarity ($S_S$).
Analogously, when two words have the same meaning (i.e., their semantic similarity is maximal, $S_S = 1$), their intelligibility index is given by their formal similarity.
Thus, we obtain the following system:

\begin{equation}
    \left\{\begin{array}{@{}l@{}}
      \alpha S_S+\beta=S_S \\
      \alpha+\beta S_L=S_L
     \end{array}\right.\,
\end{equation}

By solving the upper system, we determine the values of $\alpha$ and $\beta$ as being $\alpha=\frac{S_L(1-S_S)}{1-S_SS_L}$, respectively $\beta=\frac{S_S(1-S_L)}{1-S_SS_L}$, and, by replacing them in equation \ref{eq-li-ab} we obtain \ref{eq-li}.

The $D_{LI}$ is greater than or equal to the product of the two similarities, and it is less than or equal to the minimum between them. In other words, the following property is always true (see the proof in Appendix):
\begin{equation}
   S_SS_L\leq D_{LI} \leq min(S_S, S_L) \end{equation}

We further obtain aggregate intelligibility scores at language pair level. For each language pair $(A, B)$, where $A$ is the speaker language and $B$ is the listener language, given one of our employed corpora, we pick each sentence in language $A$ and compute an intelligibility score wrt. $B$ as the weighted average of the intelligibility indices of all the words in the sentence that are related to language $B$ (i.e. the sum of these indices divided by the total number of words in the sentence, excluding stop-words). Furthermore, we compute a corpus-level aggregate by combining all of the sentence-level scores (equivalent to combining all of the sentences into a singular text). In this way, the final overall intelligibility score between two languages will be affected both by the usage of related words in corpora in context as well as by their mutual individual intelligibility.

% Ana (before): as a weighted average of scores per language pair ? The weighted sum of word pair scores (according to their frequency in the aligned corpus) is divided by the length of the corpus

\section{Results Analysis} 
\label{sec:results}

% AICI PARCA AS SPUNE EXPERIMENT AND RESULTS ANALYSIS. AS INCEPE CU O SUBSECTIUNE NUMITA EXPERIMENT, SI AS SPUNE IN CUVINTE CE VREM SA FACEM, CHIAR DACA RELUAM PE SCURT. CA LUAM COPUSURILE SI CALCULAM INTRE ORICARE DOUA LIMBI INTELIGIBILITATEA PE BAZA METODOLOGIEI DE LA 2. CA AM FACUT PE 3 COPRUSURI, ADICA CE AVEM.

% APPOI AS PUNE O SUBSECTIUNE CU RSULTS ANAALYSIS, CAM CUM AVEM ACUM.

\begin{figure}[h!]
    \centering
    \begin{subfigure}[b]{0.35\textwidth}
        \centering
    \includegraphics[width=\textwidth]{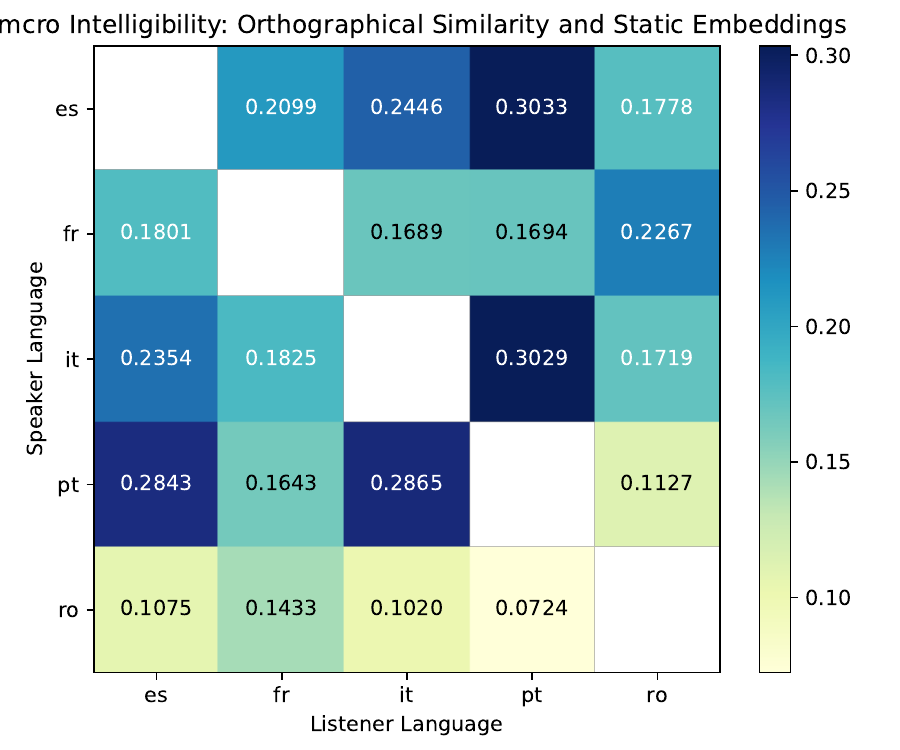}
        \caption{$D_{LI}^s$ with orthographic similarity.}
        \label{fig:heatmap-ort}
    \end{subfigure}
    \hfill
    \begin{subfigure}[b]{0.35\textwidth}
        \centering
    \includegraphics[width=\textwidth]{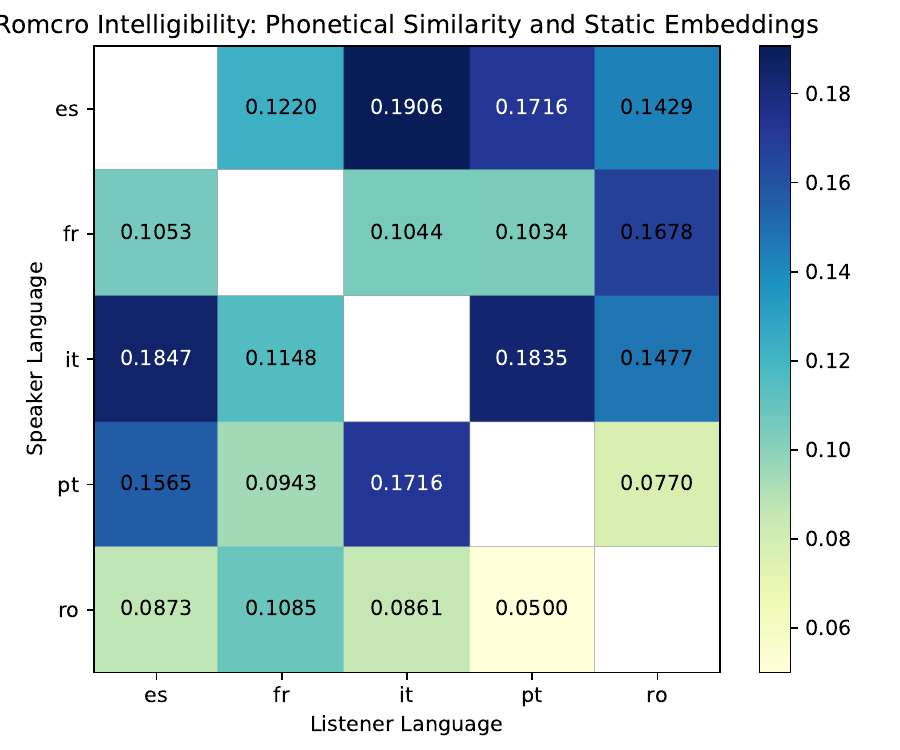}
        \caption{$D_{LI}^s$ with phonetic similarity.}
        \label{fig:heatmap-pho}
    \end{subfigure}
    \caption{Lexical intelligibility scores based on the RomCro corpus using static embeddings.}
    \label{fig:heatmaps}
\end{figure}

Table \ref{tab:results} shows resulted intelligibility scores in different settings for each language pair, computed on each of the parallel corpora, based on either orthographic or phonetic surface similarities, and using either static or contextual semantic representations for computing semantic similarities. The heatmaps in Figures \ref{fig:heatmap-ort} and \ref{fig:heatmap-pho} illustrate language intelligibility scores computed using the $D_{LI}$ index with static pretrained embeddings and the RomCro corpus. 
% Scores using the other metrics and corpus are shown in the Appendix. 

% VEDETI CA AI CI IN FIGURE SUNT ACELEASI VALORI PUSE, SUNT IDENTICE FIGURILE A SI B

The results confirm and reinforce with quantitative data the intuition we described in the introduction: it is immediately clear that for all language pairs intelligibility is \textit{asymmetrical}: there are differences between how well a speaker of language $A$ understands language $B$ (shown across the rows of the heatmaps) and how well he is understood by native speakers of language $B$ in his own language (shown on heatmap columns). For example, Spanish is understood by Portuguese speakers according to an intelligibility degree of 30.3, while Spanish speakers understand Portuguese to a degree of 28.4. These asymmetries are generally lower for core Romance languages (Spanish, Italian, Portuguese, French), and much larger for Romanian: Romanian speakers understand other Romance languages much more easily than Romanian is understood by speakers of other languages. The largest scores are obtained for Spanish and Italian, followed by Spanish and Portuguese, and the lowest for pairs involving Romanian.

Overall, the intelligibility scores based on phonetic representations are lower than those using orthographic representations, across language pairs. Interestingly, the discrepancies between orthographic and phonetic intelligibility are lowest for Romanian (a highly phonetic language). 
Generally, scores obtained using contextual embeddings, for both corpora, are lower than those using static embeddings. The discrepancies in semantic similarity scores distributions obtained using static and contextual embeddings might stem from the much better coverage of our related words vocabulary in the case of static embeddings, which are trained on a larger corpus (Wikipedia) than the two parallel corpora used to extract the contextual representations (the Appendix shows coverage scores for words in our vocabulary across the corpora). Moreover, we suggest that, while RomCro and Europarl have the advantage of being strictly parallel corpora, Wikipedia is still superior due to its size and leads to more robust semantic representations. Some differences between results across the two corpora show that domain specificity is still relevant when considering intelligibility. The versions of the index using embeddings trained on Wikipedia tend to diminish this effect and might constitute a more accurate reflection of layman comprehension. 
% In the following, we focus on $D_{LI}$ results obtained using static embeddings, computed on the RomCro corpus, which is more representative to general spoken language than the more specialized Europarl.

% ?Static embeddings might also be more suitable for reflecting semantic intelligibility since they essentially encapsulate larger contexts than BERT-based models which are inherently limited by context size (upper bounded by 512 tokens in the case of our model, but practically limited to shorter values given by sentence lengths in the corpus) - speakers trying to understand a foreign language might rely on more general notions of word meaning, learned ... - this is confirmed by results in human cloze tests, which better correlate with $D_{LI}$ intelligibility scores based on static embeddings rather than contextual ones.  (??)

% We compare our proposed metric with baselines consisting of the standalone similarities (orthographic, embedding-based and phonetic), as well as their simple product.

\begin{table*}[ht]
\centering
\small
\begin{adjustbox}{max width=\textwidth}
\begin{tabular}{lcccccccc}
\toprule
\multirow{3}{*}{\textbf{Lang. Pair}} 
& \multicolumn{4}{c}{\textbf{RomCro}} & \multicolumn{4}{c}{\textbf{EuroParl}} \\
\cmidrule(lr){2-5} \cmidrule(lr){6-9}
 & \multicolumn{2}{c}{Orthographic} & \multicolumn{2}{c}{Phonetic} 
 & \multicolumn{2}{c}{Orthographic} & \multicolumn{2}{c}{Phonetic} \\
\cmidrule(lr){2-3} \cmidrule(lr){4-5} \cmidrule(lr){6-7} \cmidrule(lr){8-9}
 & $D_{LI}^s$ & $D_{LI}^c$ & $D_{LI}^s$ & $D_{LI}^c$ 
 & $D_{LI}^s$ & $D_{LI}^c$ & $D_{LI}^s$ & $D_{LI}^c$ \\
\midrule
it-es & 23.5 & 15.4 & 18.4 & 12.1 & 21.6 & 14.6 & 16.5 & 11.0 \\
es-it & 24.4 & 14.7 & 19.0 & 11.2 & 23.1 & 15.2 & 17.6 & 11.3 \\
it-fr & 18.2 & 11.2 & 11.4 & 6.5 & 16.5 & 10.3 & 10.2 & 6.1 \\
fr-it & 16.8 & 10.3 & 10.4 & 5.9 & 14.9 & 9.6 & 9.3 & 5.6 \\
it-pt & 30.2 & 24.0 & 18.3 & 13.7 & 34.9 & 29.0 & 20.2 & 15.6 \\
pt-it & 28.6 & 22.5 & 17.1 & 12.7 & 33.1 & 27.8 & 19.1 & 14.8 \\
it-ro & 17.1 & 15.5 & 14.7 & 12.8 & 18.5 & 14.7 & 16.2 & 12.6 \\
ro-it & 10.1 & 9.2 & 8.6 & 7.5 & 14.8 & 11.7 & 12.9 & 10.1 \\
es-fr & 20.9 & 14.6 & 12.2 & 8.1 & 22.4 & 17.5 & 13.6 & 10.3 \\
fr-es & 18.0 & 13.1 & 10.5 & 7.2 & 18.7 & 15.0 & 11.5 & 8.9 \\
es-pt & 30.3 & 22.6 & 17.1 & 11.8 & 27.3 & 19.3 & 15.7 & 10.7 \\
pt-es & 28.4 & 22.3 & 15.6 & 11.6 & 25.6 & 19.5 & 14.6 & 10.8 \\
es-ro & 17.7 & 16.7 & 14.2 & 13.0 & 23.0 & 22.7 & 19.1 & 18.3 \\
ro-es & 10.7 & 10.5 & 8.7 & 8.3 & 16.7 & 17.0 & 14.0 & 13.7 \\
fr-pt & 16.9 & 12.7 & 10.3 & 7.2 & 15.7 & 12.3 & 9.2 & 7.0 \\
pt-fr & 16.4 & 12.6 & 9.4 & 6.9 & 15.9 & 11.3 & 8.6 & 6.0 \\
fr-ro & 22.6 & 20.6 & 16.7 & 14.0 & 32.4 & 30.2 & 24.7 & 21.8 \\
ro-fr & 14.3 & 13.7 & 10.8 & 9.8 & 27.4 & 24.5 & 21.2 & 18.3 \\
pt-ro & 11.2 & 9.4 & 7.7 & 6.0 & 11.3 & 9.3 & 7.7 & 5.9 \\
ro-pt & 7.2 & 6.1 & 5.0 & 3.9 & 9.5 & 8.1 & 6.6 & 5.3 \\
\bottomrule
\end{tabular}
\end{adjustbox}
\caption{Results (in \%): $D_{LI}$ intelligibility index using static ($D_{LI}^s$) and contextual ($D_{LI}^c$) embeddings for the two corpora. We also separate the computation of the index into orthographic and phonetic, depending on what strings we compute the Levenshtein distance on.}
    \label{tab:results}
\end{table*}

 % EU PARCA AS PUNE SUBSECTIUNEA ASTA DISCUSSION SI URMATOAREA INTR-O SECTIUNE SEPARATA, CU CEVA rESULTS ANALYSIS POATE AICI, SAU O DENUMIRE ASEMANATAORE CARE SA LE CUPRINDA PE AMNDOUA. PAR FOARTE BUNE AMBELE SI SE PIERD AICI 
\subsection{Discussion}
\label{sec:analysis}

% A first striking observation is that in all language pairs, the degree of intelligibilty based on phonetic forms is lower than the one based on their ortographic variants, regardless of the corpus. 
The lower degree of intelligibility scores based on phonetic forms compared to the ones based on their orthographic variants highlights, on the one hand, that in literate societies accustomed to written text, the flow of speech is more difficult to follow and understand than a text; in other words, sight has gained greater importance than hearing in recent European centuries.\todo{Anca: oare, sau pur si simplu limba vorbita se schimba mai repedede decat cea scrisa? e cam speculativ} \todo{Era gresita figura heatmap cu similaritati fonetice, poate merg revazute comentariile}
On the other hand, it recalls that writing is conservative, which brings languages closer together, as it relates them to a period closer in time to Proto-Romance; in some languages, such as French, pronunciation has evolved so much from spelling — which conforms the pronunciation of several centuries ago — that when hearing an oral discourse, a speaker of another Romance language who has no previous knowledge of French cannot spontaneously relate the vocabulary they hear to that of their own language, as pronunciation distorts it, often making it unrecognizable. In addition, the occurrence of sounds that are particular to each language - or, in any case, that are not present in both languages of a pair - (such as Fr. /\~{a}/, /\~\oe/; Pt. /\textturna/, /\textbari/, or Ro. /ts/, /d\textyogh/, etc.) pose a considerable obstacle to intelligibility. % -- DE REPARAT CARACTERELE SPECIALE... Bogdan: le-am reparat folosind https://www.johndcook.com/unicode_latex.html si biblioteca "tipa"

% When analyzing a pair of cognates $w_1$ (from language $L_1$) - $w_2$ (from $L_2$) the algorith/ also provides us the word from L2 that is semantically closest to w1 and vice versa, both obtained through embeddings.
One important source of discrepancies in intelligibility stems from semantic divergences in cognate words and borrowings across languages.  To give an example, the pair of cognates It. \textit{preparare} "prepare, get (someone) ready, cook, etc."– Ro. \textit{prepara} "prepare" only partially overlap semantically, as the conceptual area of the Romanian word is limited to preparing chemicals, medicines, a meaning limited to science; for the It. phrase "preparare la cena" (to cook dinner) the Ro. equivalent would not be "a prepara cina", but "a pregăti cina". Therefore the most similar lexeme from a semantic point of view for It. \textit{preparare} is Ro. \textit{pregăti}. The quantitative relationships extracted from Europarl are as follows: It. \textit{preparare} has 1,958 occurrences – Ro. \textit{prepara} has 44 occurrences: their usage coincides in parallel sentences in 29 instances; Ro. \textit{pregati} with 3,150 occurrences – coincides with It. \textit{preparare} in 1,167 cases; the semantically closest lexeme for Ro. \textit{prepara} still is It. \textit{preparare}, as it also applies perfectly in the scientific area covered by the Romanian word. 
One case in which the polysemy of words makes it difficult to identify a perfect equivalent in discourse (not in lexicography) is the pair It. \textit{oscuro} "dark, obscure" – Ro. \textit{obscur} "unknown, dark (= evil)". In the RomCro corpus, It. \textit{oscuro} has 490 occurrences, and Ro. \textit{obscur} 113 (overlapping in 87 cases); the closest word to Ro. \textit{obscur} is It. \textit{misconosciuto} ("un personaj obscur" = an unknown person), which has 6 occurrences in RomCro, but never coincides with It \textit{oscuro}; the closest to It. \textit{oscuro} in (static) embedding space is Ro. \textit{malefic} "maleficent" (29 contexts, but they overlap only once).

% , according to which the degree of intelligibility is asymmetrical. They reflect the following asymmetries:
% - The closest proximity in both directions and the most balanced relationship is between sp – pt (sp-pt 28 procente, pt-sp 26), followed by it-pt (it-pt 27, pt-it 25).
% - the smallest proximity is found between ro-pt (ro-pt 06, pt-ro 09). 
The asymmetries in intelligibility are maintained across the two corpora, but the scores change slightly in the direction of increasing mutual intelligibility (without changing the asymmetry ratio) when we take Europarl as a corpus, given the specific nature of these texts, which contain specialized language from the political-economic field, language leveled by the presence of internationally widespread neologisms. However, a significant difference is represented by the relationship between Romanian and French, on the one hand by the reduction of the asymmetry gap (7 points difference in RomCro and 4 in Europarl), and on the other hand by the significant increase in the degree of intelligibility of Romanian from the perspective of the French speaker to 24.5, which represents the highest level of intelligibility for French as a target language. The explanation lies in the relationship between Romanian and French: in the 19th century, when Romanian culture underwent a period of development and French culture was its model, Romanian was exposed to a process of massive borrowing from French, with French-origin vocabulary accounting for more than 9 percent of the total Romanian lexicon. This borrowed vocabulary corresponds to literary, scientific, and legal-administrative language, which means that its occurrence in the political and economic discourse present in the Europarl corpus has a significant weight and, therefore, is highly intelligible to French speakers.

The result concerning the level of mutual intelligibility between Portuguese and Romanian requires a reconsideration of the lateral areas theory \cite{bartoli1925introduzione}, which claimed that the languages on the edge of the former Roman Empire retained many common features and differed less from each other than they did from the central languages. However, this study shows that geographically peripheral languages have not remained as similar as previously thought, but have, on the contrary, diverged. Direct contact proves to be more important than common heritage and conservative tendencies: thus, the intelligibility between Spanish and Portuguese is among the highest, and the asymmetry is also much smaller, as the Iberians evolved in constant contact with each other. 

The highest degree of asymmetry in mutual intelligibility occurs in pairs that have Romanian as one of the members.
% : the difference between levels is, in all cases (except for Ro-Pt), 6 points, unlike the level of asymmetry in the other pairs, which does not exceed 2 points.
% So, the question is: why do Romanian speakers understand Romance languages much better than they are understood? 
An illustrative example of how this occurs is the following case: all Romance languages except Romanian have inherited the Latin word \textit{mundus} "world" (Es \textit{mundo}, Pt \textit{mundo}, Fr \textit{monde}, It \textit{mondo}). Romanian uses the word \textit{lume} for this concept, inherited from Lat. \textit{lumen} "light", whose cognates in other Romance languages, if they exist, are too different phonetically and/or semantically to allow for correct identification (e.g. Es. \textit{lumbre} "light," It. \textit{lume} "candel"). On the other hand, Romanian has borrowed the French adjective \textit{mondial}, "worldwide", which allows Romanian speakers to understand Romance words with the same root, while the speakers of other Romance languages have no clue for understanding Ro. \textit{lume} "world".
This category of words, inherited in all Romance languages except Romanian, is fairly well represented \cite{fischer1964panroman}
% (Fischer 1964 - Fischer, I., Panroman sauf roumain, Revue roumaine de linguistique, vol.9, Nr.6, 1964, p.595-602) 
and contributes to the asymmetry observed. Another factor that leads to the reduced intelligibility of Romanian is the Slavic component of the vocabulary, which amounts to almost 10\% and covers basic concepts that in other languages are verbalized by words of Latin origin (e.g., Ro. \textit{sărac} "poor”, \textit{boală} "disease", \textit{dușman} "enemy", etc.).
% TODO: de reparatdiacriticele

\subsection{Comparison to Human Intelligibility Tests}
\label{sec:evaluation}

\begin{figure}
    \centering
    \includegraphics[width=0.85\linewidth]{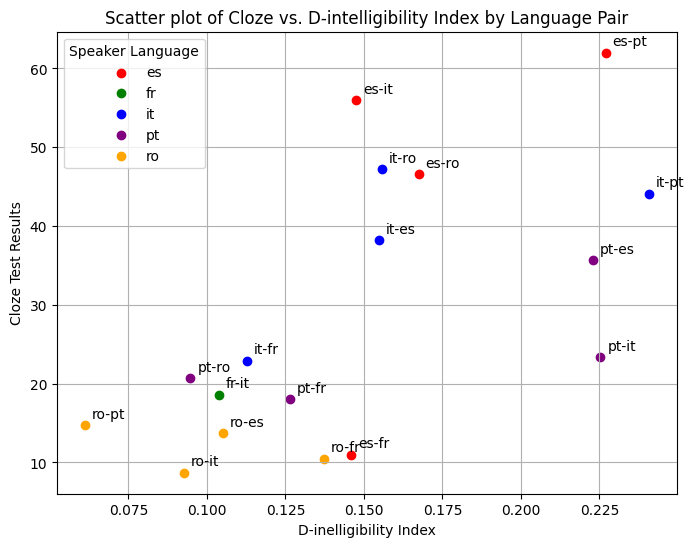}
    \caption{$D_{LI}^s$ index (based on RomCro) and cloze test results, for each language pair $A-B$ ($A$ as speaker language and $B$ as listener language), colored according to speaker's language.}
    \label{fig:scatter-corr}
\end{figure}

Language similarity measures are inherently difficult to evaluate, since there is no gold standard to refer to. We thus propose a computational metric of language intelligibility and attempt to understand how it compares to intelligibility in humans by resorting to results of human subject experiments in a cloze test. Such an experiment applied to Romance language speakers in each of the other Romance languages is reported in \citet{gooskens2018mutual}, which we take as an estimate of human intelligibility across the Romance languages.
Since cloze test results (measuring the accuracy of participants in filling in the correct word in context) are not directly comparable to our metric, we compare the rankings of intelligibility scores for each language pair, using Spearman correlation. We find that cloze test results are significantly correlated with our $D_{LI}$ index, with a correlation coefficient of 0.71 (p-value 0.0013) based on RomCro (roughly identical using either or static embeddings).
% confirming that interestingly (?) the correlation is higher for our D-index than for the simple embedding similarity, with either static or contextual embeddings.
Figure \ref{fig:scatter-corr} shows a comparison between cloze test results and $D_{LI}^s$ scores for each language pair.
For the $D_{LI}^s$ index based on Europarl the correlation only reaches 0.44 (with p-value 0.07, while $D_{LI}^c$ is not significantly correlated), which might be explained by the more specialized vocabulary, unlike the language test texts used in the cloze test. Correlations using phonetic distances are similar (correlation up to 0.76 with static embeddings on RomCro). Differences might be related to the nature of the cloze test (based on a combination of listening and choosing missing written words).
% [TODO Comparatia grafic cu fonetic? Fonetic sunt mari corelatiile - Appendix?]

% It is noteworthy that the relationships between languages and the degrees of asymmetry obtained by us strictly through lexical analysis correlate very well with the results obtained by \citet{gooskens2024mutual} through cloze tests. 
% |De exemplu: (de facut un mic tabel?)
% Gooskens (2024, 228)
% - francezul înțelege it. 24,2, sp. 31,5, ro. 11,0
% - nouă (Romcro) ne-a ieșit că francezul înțelege it. 15,5, sp. 18, ro. 12,3
% Gosskens: it. înțelege fr. 46,3, pt. 33,5, ro. 10,6, sp. 65,7
% - la noi: it intelege fr 14,2, pt 25,5, ro 8,6 si sp. 21,8
% Gooskens: ro înțelege fr. 47,1, it. 57,7, dar ptg. doar 22,9
% - la noi ro intelege fr 19,6, it. 14,8, pt. 9,2

% In both analyses, Ro measures the lowest levels of intelligibility from the perspective of all languages. In the analysis provided by \citet{gooskens2024mutual}, this level does not exceed 15 percent, and in our analysis, the maximum level is 12.3 percent (for Fr.). 

% TODO: de comparat cu cloze testul facut de noi la false friends. Cov1 pe mask, ala ar fi un cloze test facut de BERT

It should be noted that the comparison is limited in several ways. First of all, the data used in the cloze test experiments is different from the data in our experiments. The variation between the correlations for the two corpora proves the importance of the corpus. Secondly, the corpus used in the cloze test is very limited, consisting of 4 texts of $\approx$200 words each, with only a few target words per text, which are unlikely to cover a representative portion of the vocabulary of related words (including all variations of cases of semantic divergence) that are essential to how intelligibility functions.
We thus use the comparison to the human intelligibility score as an indication of our metric's capability to reflect to some extent the way that human intelligibility functions, without relying on it as an absolute gold standard. 
% One of the advantages of our metric is precisely that it can be used on larger corpora without the need to perform human tests.
% SA ZICEM ASAT COMENTATA, CA UN AVANTAJ AL METODEI NOASTRE E CA NU AVEM NEVOIE DE SUBIECTI UMANI, CA FACEM AUTOMAT

\section{Conclusions}
\label{sec:conclusion}
In this paper we presented an analysis of intelligibility across related languages with a focus on the five main Romance languages. We propose to quantify mutual intelligibility in related languages using computational tools, based on etymologically related words and their usage in corpora as captured by modern word embeddings. We introduced a novel lexical intelligibility index which we aggregate to obtain a map of intelligibility scores across the five Romance languages, including orthographic, phonetic, and semantic information, based on two different parallel corpora, confirming intuitions about related languages intelligibility and its asymmetry. In the future, we propose that spoken language corpora might improve the reliability of the computed intelligibility scores.

\section*{Limitations}

Evaluating intelligibility scores in comparison to results on human experiments are limited by the size and nature of the texts used in these experiments. We propose that using automatic metrics of intelligibility such as the one proposed in this paper is one way to overcome this problem and leverage larger corpora which can be exploited automatically more efficiently.
Nevertheless, even for computational measures of intelligibility, there are still limits in the availability of large enough corpora, especially parallel corpora. We use some of the largest available parallel corpora, but they are still limited in size (much smaller than Wikipedia for example, leading to lower coverage of related words and lower quality of embeddings) and domain (literary texts and political speeches). The low coverage of related words in the embeddings spaces also means that, in the case of contextual embeddings, it was necessary to aggregate contextual representations across all contexts found in the corpus in order to obtain robust representations. Larger parallel corpora or multilingual models might allow us to better compare individual parallel contexts without having to construct global representations.

% - parallel corpora in size, especially spoken language corpora. Future: find spoken language corpora
% - evaluation limited by human experiments and by the difficulty of the question - we propose that this is an argument for
% - computational limitations in extracting embeddings on very large corpora limit how much we can rely on contextual embeddings based on large corpora such as wikipedia. The superiority of the smaller static models in our experiments 

Some refinements of the data in the lexicons used, such as excluding words with multiple etymologies, which are not currently handled in the available version of the database, could be useful for a more accurate model of the linguistic phenomenon.  
Graphic issues can also lead to errors in detecting cognates, such as some spelling errors related to diacritics. In this study, we rely on stemming and heavy normalization to account for this, but corrections directly in the related word database would render this unnecessary and lead to more precise matching.

% The Italian term \textit{rio} (‘river’) and the Romanian term \textit{rău }(‘bad’) were identified as false friends. However, the two terms don’t share a common etymon – the first one comes from the Latin word \textit{rivus} and the second one from lat. \textit{reus – }and as such they can be considered chance false friends. A possible explanation for this pair resides in the graphical resemblance of \textit{rău }(‘bad’) and \textit{r}\textit{âu} (‘river’) and the misidentification of the diacritics on the letter \textit{a,} \textit{ă }and \textit{â. }

\section*{Ethical Statement}
There are no ethical issues that could result from
the publication of our work. Our experiments comply with all license agreements of the data sources
used.

% \section*{Acknowledgments}

% Acknowledgments.

\bibliography{custom}

\appendix

\section{Appendix}
\label{sec:appendix}
\subsection{Intelligibility Index Properties}
The following properties show the range of $D_{LI}$ index:
\begin{equation}
   S_sS_L\leq D_{LI} \leq min(S_S, S_L) 
   \label{eq:dli-range}
\end{equation}

To prove\ref{eq:dli-range}, we proceed as follows.
To show the first part of (5), that
$$S_SS_L\leq D_{LI}=\frac{S_S S_L(2-S_S -S_L)}{1-S_SS_L}$$, this is equivalent to show that $$1-S_SS_L \leq 2-S_S -S_L$$ (because both $S_L$ and $S_S$ are positives), which is equivalent to showing that $$S_S - S_SS_L\leq 1-S_L$$, which is equivalent to showing that $$S_S(1-S_L)\leq 1-S_L$$. But since $$S_L\leq 1 $$, we can simplify by $1-S_L$ and the last equation is equivalent to $S_S \leq 1$, which is always true.

For the second half of the inequality \ref{eq:dli-range} , we will show that $D_{LI} \leq S_S$ (to show that ,$D_{LI} \leq S_L$ is analogous)
$D_{LI} \leq S_S$ is equivalent to $$\frac{S_S S_L(2-S_S -S_L)}{1-S_SS_L}\leq S_S$$, which is equivalent to showing that $$\frac{S_L(2-S_S -S_L)}{1-S_SS_L}\leq 1$$ (because $0 \leq S_S$), which is equivalent to $$2S_L-S_S S_L -S_LS_L \leq 1-S_SS_L$$, which is equivalent to $$0 \leq 1 -2S_L +S_L^2=(1-S_L)^2$$, which is always true.

\subsection{Additional Results: Word Coverage and Score Distributions}

\begin{figure*}[h!]
    \centering
    \begin{subfigure}[b]{0.45\textwidth}
        \centering
        \includegraphics[width=0.9\textwidth]{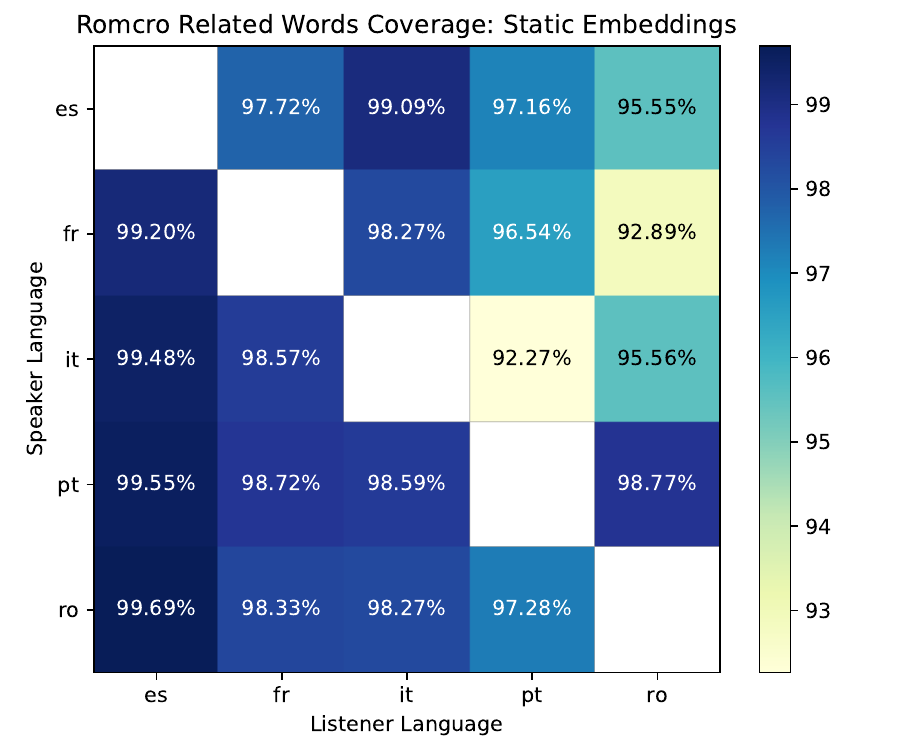}
        \caption{Coverage of static embedding space trained on Wikipedia for related words in the RomCro corpus.}
        \label{fig:sub1a}
    \end{subfigure}
    \hfill
    \begin{subfigure}[b]{0.45\textwidth}
        \centering
        \includegraphics[width=0.9\textwidth]{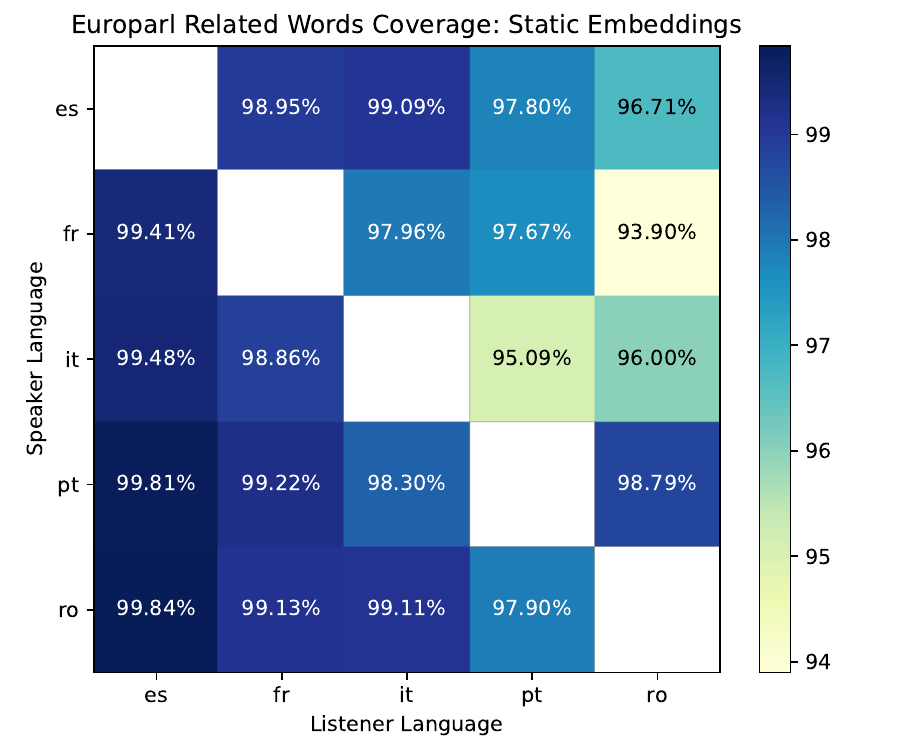}
        \caption{Coverage of static embedding space trained on Wikipedia for related words in the Europarl corpus.}
        \label{fig:sub2a}
    \end{subfigure}
    \quad % Add some horizontal space between the subfigures in the next row if needed, or just start a new line
    \begin{subfigure}[b]{0.45\textwidth}
        \centering
        \includegraphics[width=0.9\textwidth]{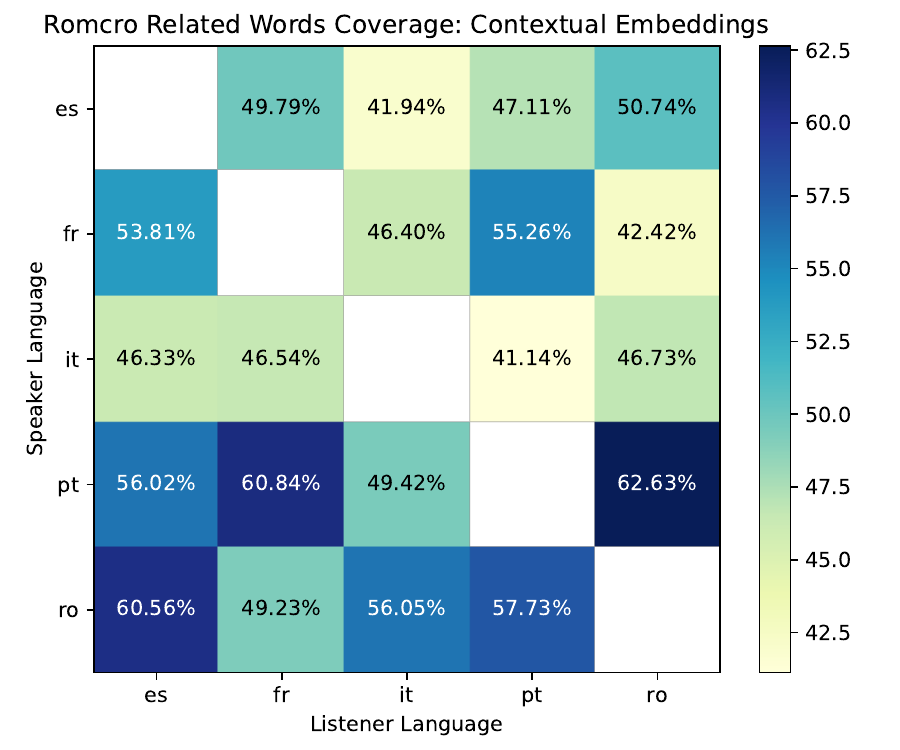}
        \caption{Coverage in contextual embedding space for related words for the RomCro corpus.}
        \label{fig:sub3}
    \end{subfigure}
    \hfill
    \begin{subfigure}[b]{0.45\textwidth}
        \centering
        \includegraphics[width=0.9\textwidth]{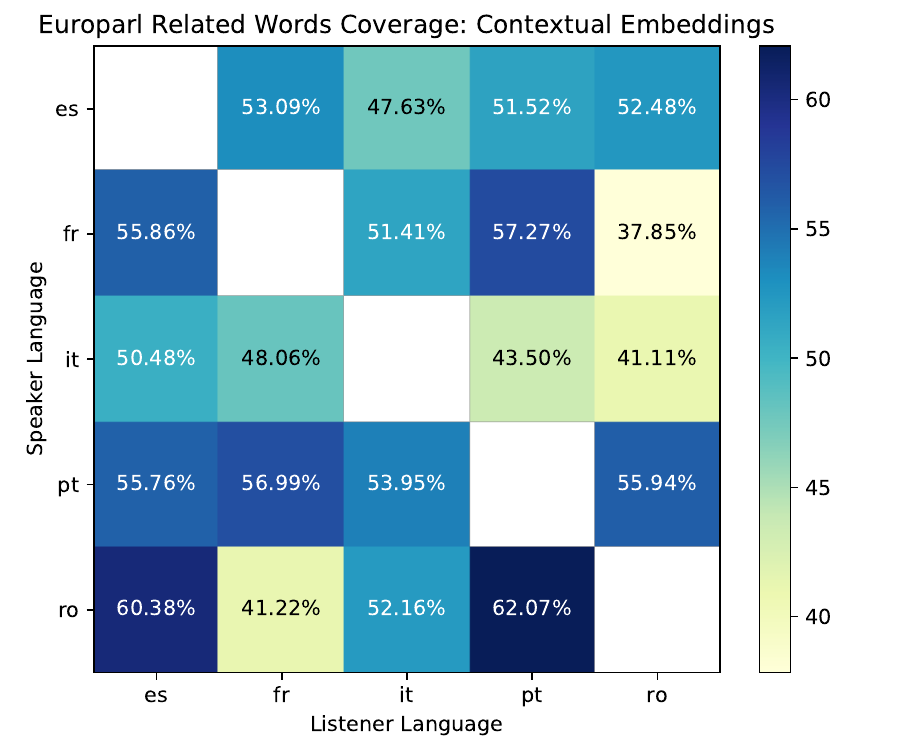}
        \caption{Coverage in contextual embedding space for related words for the Europarl corpus.}
        \label{fig:sub4}
    \end{subfigure}
    \caption{Embeddings coverage.}
    \label{fig:main}
\end{figure*}

\begin{figure*}
    % \centering
    \includegraphics[width=\linewidth]{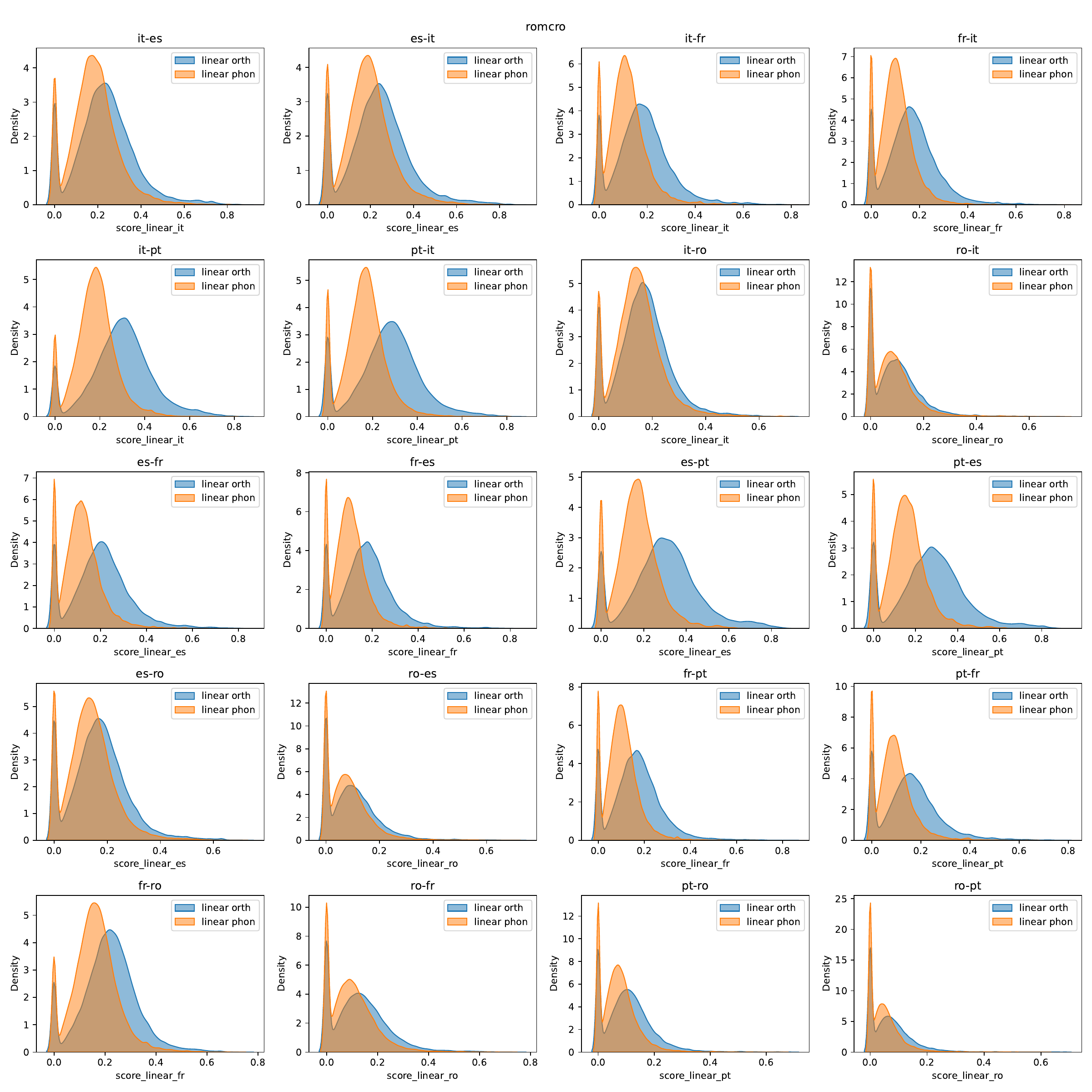}
    \caption{Distribution of $D_{LI}$ scores using static embeddings and orthographic vs phonetic surface similarity.}
    \label{fig:sub1b}
\end{figure*}

\begin{figure*}
    % \centering
    \includegraphics[width=\linewidth]{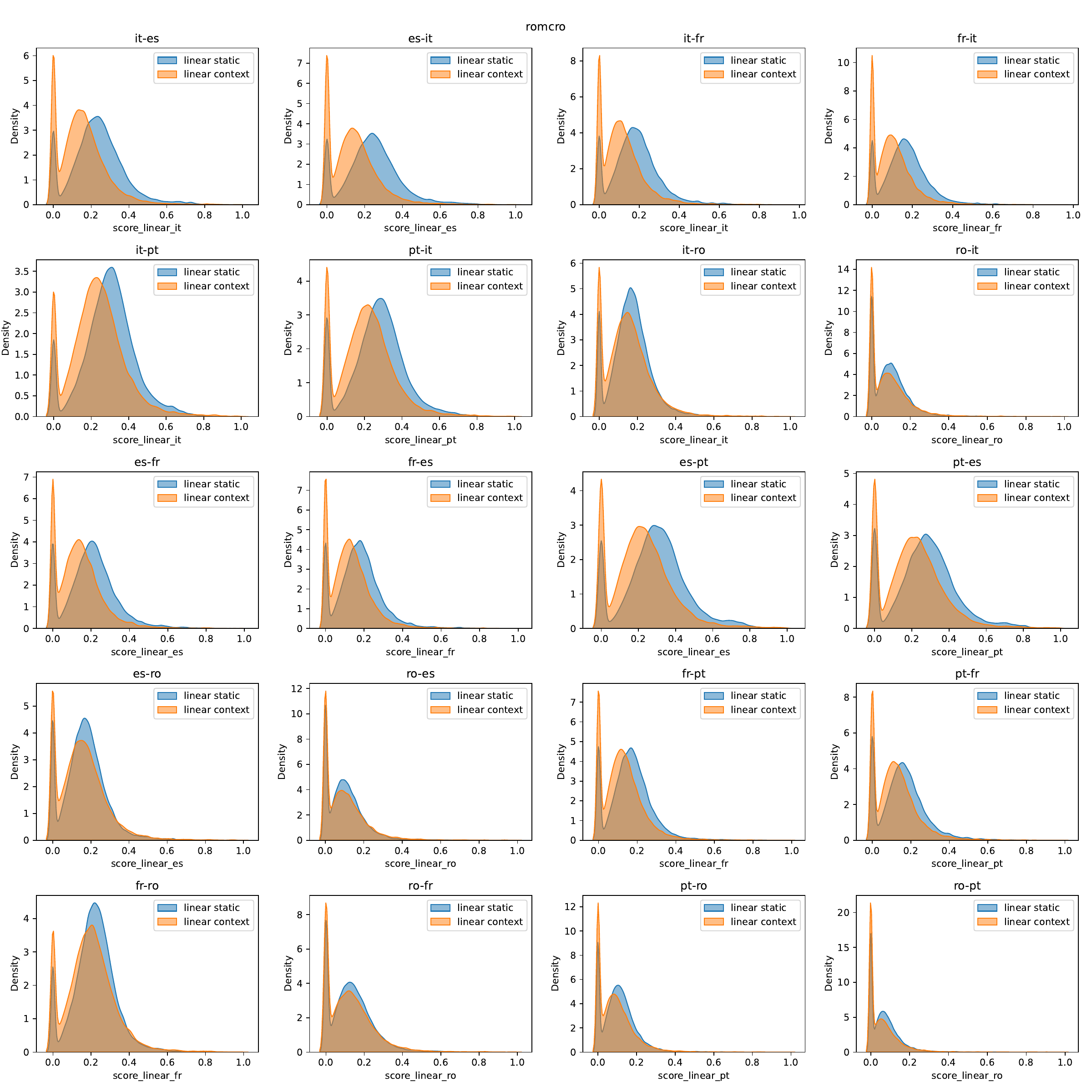}
    \caption{Distribution of $D_{LI}$ scores using orthographic similarity and static vs contextual embeddings for semantic similarity.}
    \label{fig:sub2b}
\end{figure*}

% \caption{Intelligibility scores distributions across word pairs for each language pair using the RomCro corpus.}
% \label{fig:main}

\begin{figure*}
    % \centering
    \includegraphics[width=\linewidth]{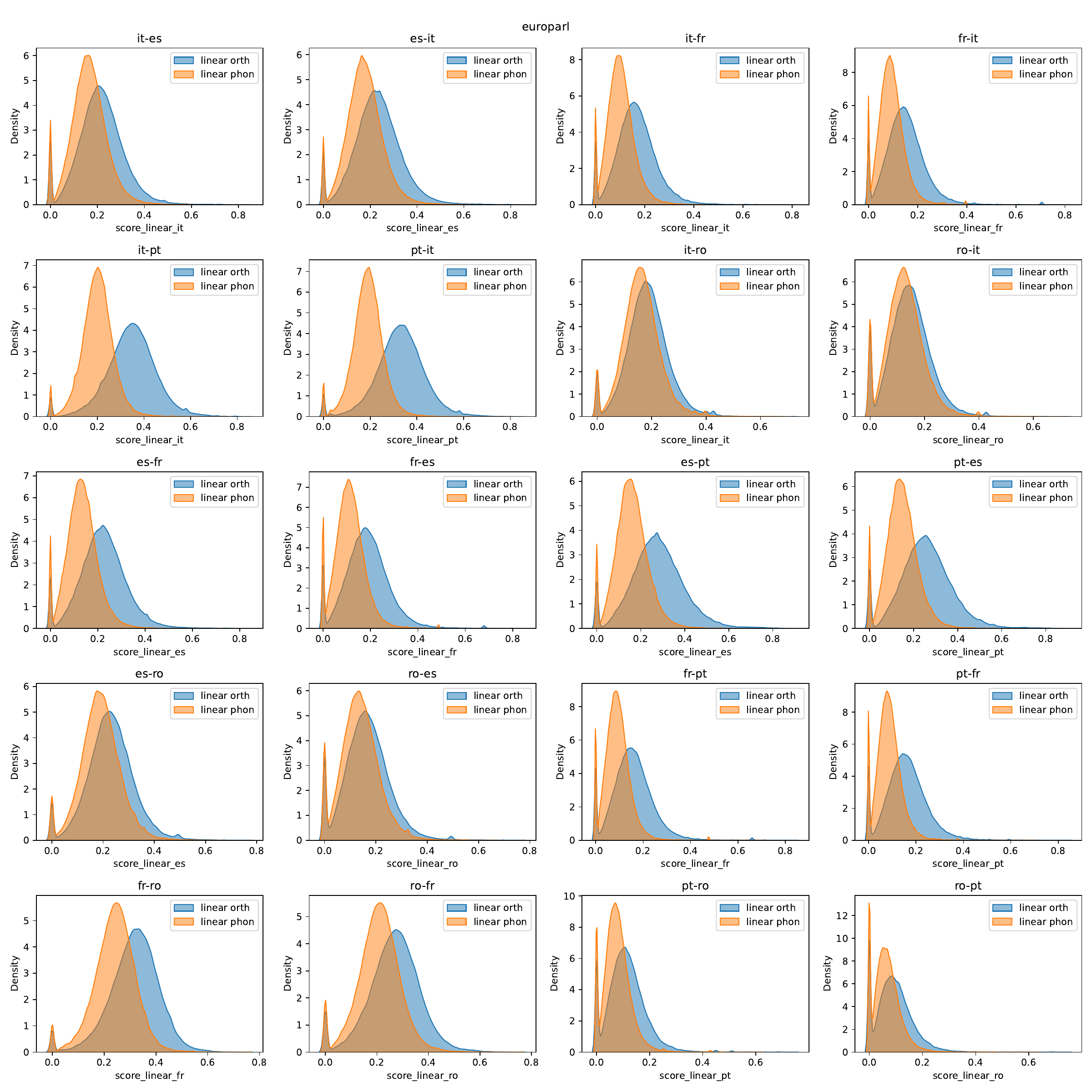}
    \caption{Distribution of $D_{LI}$ scores using static embeddings and orthographic vs phonetic surface similarity.}
    \label{fig:sub1c}
\end{figure*}

\begin{figure*}
    % \centering
    \includegraphics[width=\linewidth]{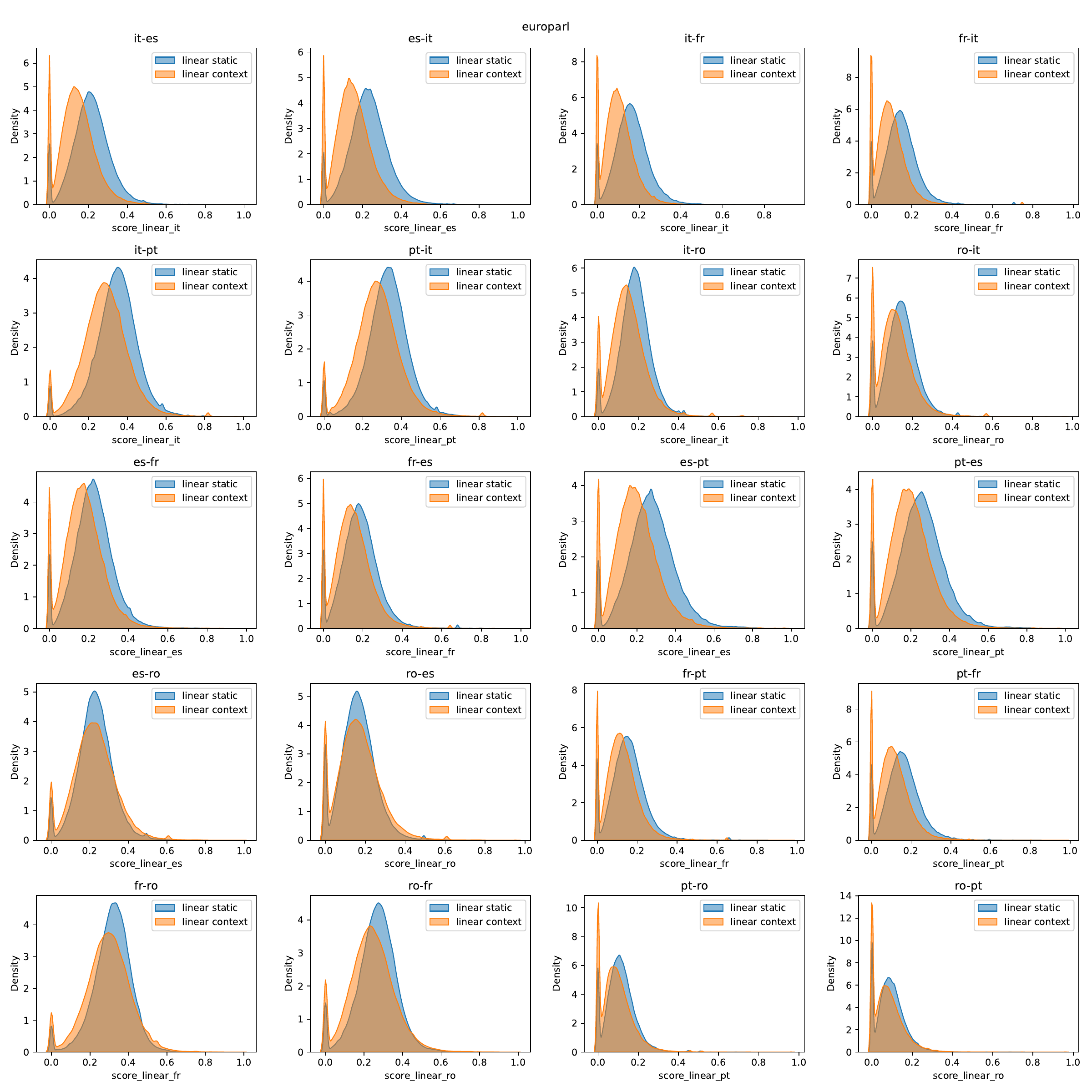}
    \caption{Distribution of $D_{LI}$ scores using orthographic similarity and static vs contextual embeddings for semantic similarity.}
    \label{fig:sub2c}
\end{figure*}

% \caption{Intelligibility scores distributions across word pairs for each language pair using the Europarl corpus.}
% \label{fig:main}

\subsection{Infrastructure and libraries}

The experiments were performed on an RTX 2080 Ti GPU and a Ryzen 5 3600X CPU for a total of 72 hours.

Libraries used for embedding extraction, cognate and corpora preprocessing (extracting stems), synonym extraction based on WordNet, and distance metrics computation:
\begin{itemize}
    \item keras==3.8.0
    \item keras-hub==0.18.1
    \item keras-nlp==0.18.1 
    \item nltk==3.9.1 
    \item scikit-learn==1.6.1
    \item scipy==1.13.1 
    \item sentence-transformers==3.4.1
    \item spacy==3.7.5
    \item tensorflow==2.18.0
    \item tensorflow-datasets==4.9.7 
    \item transformers==4.48.3
    \item and fasttext vector support based on \url{https://github.com/babylonhealth/fastText_multilingual/.}
\end{itemize}

Transformer models used:
\begin{itemize}
    \item \texttt{distiluse-base-multilingual-cased-v2}: 135M parameters
    % \item \texttt{xlm-roberta-base}: 279M parameters
\end{itemize}

Hyperparameters:
\begin{itemize}
    \item maximum number of sampled occurrences for a word when computing contextual embeddings: 200
    \item occurrence matching was checked based on stem matching with and without unicode normalization (removing of accents)
    \item Affinity Propagation clustering was trained with the default hyperparameters provided by the scikit-learn library.
\end{itemize}

\end{document}